\documentclass[10pt, twocolumn]{article}
\usepackage[a4paper, total={7in, 9in}]{geometry}

\usepackage[numbers]{natbib}
\usepackage{hyperref}
\usepackage{amssymb}
\usepackage{latexsym}
\usepackage{amsmath}
\usepackage{amssymb}
\usepackage{amsthm}
\usepackage{xcolor}
\usepackage{comment}
\usepackage{subcaption}
\usepackage{tikz}
\usepackage{pgfplots}
\usepackage{algorithm}
\usepackage{algpseudocode}
\usepackage{cleveref}
\usepackage{url}
\usepackage{lipsum}
\usepackage{array}
\usepackage{comment}
\usepackage{subcaption}
\usepackage{etoolbox}
\usepackage{wasysym}

\usepackage{url}
\usepackage{xcolor}
\definecolor{newcolor}{rgb}{.8,.349,.1}

\DeclareMathOperator*{\prunetop}{Top}
\DeclareMathOperator*{\prunegreedy}{GLP}
\DeclareMathOperator*{\pruneopt}{Optimal}
\DeclareMathOperator*{\argmax}{argmax}

\newtheorem{assumption}{Assumption}[section]
\crefname{assumption}{assumption}{assumptions}
\Crefname{assumption}{Assumption}{Assumptions}

\newcommand{\david}[1]{\emph{\textbf{(David: #1)}}}
\newcommand{\sebastian}[1]{\emph{\textbf{(Sebastian:  #1)}}}
\newcommand{\antonio}[1]{\emph{\textbf{(Antonio: #1)}}}
\newcommand{\todo}[1]{\emph{\textbf{ToDo: #1}}}

\newcommand\citeauthorn[1]{\citeauthor{#1} \cite{#1}}

\renewcommand{\david}[1]{}
\renewcommand{\sebastian}[1]{}
\renewcommand{\antonio}[1]{}
\renewcommand{\todo}[1]{}

\begin{document}
 \setcounter{page}{1}

\title{Greedy-layer Pruning: Speeding up Transformer Models for Natural Language Processing \footnote{Accepted at Pattern Recognition Letters (\url{https://www.journals.elsevier.com/pattern-recognition-letters})}}

\author{David Peer$^{1,2}$ \qquad
Sebastian Stabinger$^{1,2}$ \qquad
Stefan Engl$^1$ \qquad
Antonio Rodr\'{i}guez-S\'{a}nchez$^2$ \\
}

\date{%
    $^1$DeepOpinion, 6020 Innsbruck, Austria\\ %
    $^2$University of Innsbruck, 6020 Innsbruck, Austria\\[2ex]%
    \today
}

\maketitle

\begin{abstract}
\emph{
Fine-tuning transformer models after unsupervised pre-training reaches a very high performance on many different natural language processing tasks. Unfortunately, transformers suffer from long inference times which greatly increases costs in production. One possible solution is to use knowledge distillation, which solves this problem by transferring information from large teacher models to smaller student models. Knowledge distillation maintains high performance and reaches high compression rates, nevertheless, the size of the student model is fixed after pre-training and can not be changed individually for a given downstream task and use-case to reach a desired performance/speedup ratio. Another solution to reduce the size of models in a much more fine-grained and computationally cheaper fashion is to prune layers after the pre-training. The price to pay is that the performance of layer-wise pruning algorithms is not on par with state-of-the-art knowledge distillation methods. In this paper, \emph{Greedy-layer pruning} is introduced to (1) outperform current state-of-the-art for layer-wise pruning, (2) close the performance gap when compared to knowledge distillation, while (3) providing a method to adapt the model size dynamically to reach a desired performance/speedup tradeoff without the need of additional pre-training phases. Our source code is available on} \url{https://github.com/deepopinion/greedy-layer-pruning}.
\end{abstract}

\section{Introduction}\label{sec:introduction}

Fine-tuning transformer models on multiple tasks after pre-training is the de-facto standard in Natural Language Processing \cite{multitask-learning-nlp} due to its outstanding performance on a wide range of applications such as sentence classification \cite{bert, roberta}, named entity recognition \cite{transformer-named-entity-recognition} or document level relation extraction tasks \cite{document-level-relation-extraction}. Classical transformer models, such as BERT \cite{bert}, are massive networks with up to 24 layers, each containing fully connected layers as well as many attention heads, leading to more than 100M parameters as a result. This extremely large number of parameters leads to long inference times and an increase in cost if deployed in production after training. Additionally, increased memory consumption and inference-time hinder the deployment of transformer models on edge- or embedded devices. Different areas of research address this problem by targeting a reduction in memory consumption as well as inference time. One such research direction is knowledge distillation, where information is transferred from large teacher models to much smaller student models, while maintaining up to $96.8\%$ of the performance of the teacher \cite{tinybert}.

In the first step of knowledge distillation, information is transferred during a pre-training stage from large teacher models to smaller student models. Specialized teacher models can be pre-trained additionally for better performance  \cite{mobilebert}. \emph{Task-agnostic} knowledge distillation models such as DistilBERT \cite{distilbert} or MobileBERT \cite{mobilebert} can directly be fine-tuned on downstream tasks after the pre-training distillation stage. Other models, such as TinyBERT, are computationally more expensive as they use the teacher also in the fine-tuning stage. In either of those methodologies, knowledge must be distilled from large teacher models into smaller student models during the pre-training phase to produce a compressed model. The knowledge-distillation during the pre-training stage is computationally expensive and therefore it is impractical to adjust the size of student models for each use-case individually to reach a desired performance/speedup ratio. \emph{Task-specific knowledge distillation} methods such as PKD \cite{pkd} do not require pre-training distillation, but the performance is also lower compared to the aforementioned approaches as we will show later in the experimental \cref{sec:experimens}. Nevertheless, having models with different performance/speedup ratios available is crucial for deployment in production and therefore most researchers provide models of different sizes in the form of pre-trained base, large and xlarge models \cite{bert, roberta}. An even more fine-grained selection would further help in practice to adjust the performance/speedup trade-off specifically for a certain use-case and downstream-task, but it is unfortunately computationally too expensive and therefore impractical to pre-train a special model or distill a special student-model for each use-case individually. Therefore, our goal is to develop a method to compress and train models on a certain downstream task after the pre-training phase to enable a cheaper and much more fine-grained selection of the performance/speedup ratio.

\citeauthorn{sixteen-heads}, \citeauthorn{analyze-multi-head-attention} and \citeauthorn{meta-pruning-attention-heads} have already shown that many attention heads can be removed without a significant impact on performance. Unfortunately, the speed-up reached by those methods is not much bigger than $1.3 \times$ because only parts of the encoder layers are pruned, making them not sufficiently competitive w.r.t inference time against distilled models. Much higher compression rates can be achieved through layer-wise pruning as complete encoder layers are removed from the model. \citeauthorn{structured-dropout} proposes a form of structured dropout, called \emph{LayerDrop}, which is used during pre-training and fine-tuning to make models robust against removing whole layers. This solution requires an additional pre-training phase that is quite expensive to compute and as we show in the experimental section results are worse compared to knowledge distillation. \citeauthorn{poor-mans-bert} studied many different layer-wise pruning approaches for transformers that directly operate on pre-trained models. They found that the performance is best if their \emph{Top-layer pruning} algorithm is used. Models that are compressed with Top-layer pruning are already on-par with DistilBERT, but still behind the novel distillation methods such as TinyBERT or MobileBERT. Our goal is therefore to outperform current state-of-the-art methods for layer-wise pruning in order to reduce the performance gap when compared against distilled models. This goal is quite a challenging one as, if we compare the current state of the art on GLUE for distillation with MobileBERT, it reaches a score of $81.0$, while state-of-the-art layer-wise pruning reaches a score of just $79.6$ (\cref{fig:timing}), which may be the reason why layer-wise pruning has been somewhat neglected until now.

We propose \emph{Greedy-layer pruning (GLP)}, an algorithm that compresses and trains transformer models while needing only a modest budget. Our approach avoids the computationally expensive distillation stage from the teacher into the student by directly reducing the size of already pre-trained base models through pruning. First, a given pre-trained model is analysed to determine an order to prune layers while maintaining high performance. Then, individual compression rates and performance/speedup ratios can be selected with a simple lookup that can be done in $O(1)$. The proposed methodology is therefore $12 \times$ less expensive compared to knowledge distillation as will be shown in the experimental \cref{sec:experimens}. The proposed methodology outperforms state-of-the-art layer-wise pruning algorithms and is competitive with state-of-the-art distillation methods w.r.t inference time and performance which makes this approach even more practicable. Some of the compressed models that GLP produces even outperform the baseline as we will show in our experimental evaluation. Our experimental results also indicate that our GLP algorithm is empirically effective for different transformer models while pruning $50\%$ of the layers, we still maintain $95.3\%$ of the performance of a BERT model and $95.4\%$ of the performance of a RoBERTa model. Therefore, we will conclude that the models can dynamically be compressed with Greedy-layer pruning without the need of additional pre-training stages while still maintaining high performance. The contributions of this paper are:
\begin{itemize}
    \item A Greedy-layer-wise pruning algorithm is introduced to prune transformer models while maintaining high performance.
    \item We experimentally show that Greedy-layer pruning outperforms the state of the art in terms of layer-wise pruning and closes the performance gap to distillation methods.
    \item It is demonstrated for each GLUE task that it is possible to select the model size with a single lookup after GLP analyzed the pre-trained model to reach a specific performance/speedup tradeoff.
    \item Greedy-layer pruning and Top-layer pruning are compared against the optimal solution to motivate and guide future research.
\end{itemize}

This paper is structured as follows: Related work is presented in the next section. In \cref{sec:method}, layer-wise pruning is defined and Greedy-layer pruning is introduced. In the experimental \cref{sec:experimens} we compare GLP against the state of the art for layer-wise pruning and for distillation, we compare the costs for compressing models with GLP as well as state-of-the-art knowledge distillation methods and finally compare layer-wise pruning methods against the optimal solution. A discussion on limitations is presented in \cref{sec:conclusion} followed by a future work section.

\section{Related Work} \label{sec:related_work}
This work falls into the research area of compressing transformer architectures. Different areas of research address the problem to reduce memory consumption as well as inference time of transformer models. A comprehensive case-study on this topic is given by \citeauthorn{compress-transformer-case-study} who categorize this area into data quantization, knowledge distillation and pruning. The proposed method falls into the research area of pruning. One pruning strategy is to focus on attention heads. The contribution made by individual attention heads in the encoder has been extensively studied. \citeauthorn{sixteen-heads} demonstrated that 16 attention-heads can be pruned down to one attention head, achieving an inference speedup of up to $17.5\%$ if the batch size is large. Another method to prune unnecessary attention heads is through single-shot meta-pruning \cite{meta-pruning-attention-heads}, where pruning $50\%$ of the attention heads, reaches a speedup of about $1.3 \times$ on BERT, which is still not enough to be competitive against distilled models. Instead of pruning only attention heads, it is also possible to prune layers of the encoder unit including all attention heads and the feedforward layer. \citeauthorn{structured-dropout} pre-trains BERT models with random layer dropping such that pruning becomes more robust during the fine-tuning stage. On the other hand, \citeauthorn{poor-mans-bert} claims that layer dropping \cite{structured-dropout} is (1) expensive because it requires a pre-training from scratch of each model and (2) it is possible to outperform this methodology by dropping top-layers before fine-tuning a model. \citeauthorn{poor-mans-bert} evaluated six different encoder unit pruning strategies, namely, top-layer dropping, even alternate dropping, odd alternate dropping, symmetric dropping, bottom-layer dropping and contribution-based dropping. They found that (1) pruning top layers from the model consistently works best, (2) layers should be dropped directly after pre-training and (3) that an iterative pruning of layers does not improve the performance. Nevertheless, Top-layer pruning is not on par with distillation methods, something that we address in this paper.

\section{Methods} \label{sec:method}
The layer-wise pruning task and the optimization problem that pruning algorithms should solve is defined as follows: Assume that a pre-trained transformer model of depth $d$ consists of $L$ layers with $|L|=d$. The layer $L_0$ is the input layer and layer $L_{d-1}$ is the layer just before the classifier. Our goal is to develop an algorithm $\mathcal{A}(L, T, n)$ for a given fine-tuning task $T$ (e.g. QNLI) that returns a subset $R_n \subset L$ with $|R_n| = n$. We call an algorithm $\mathcal{A}$ \emph{task-independent} if it returns the same solution $R_n$ for all tasks $T$ and \emph{task-dependent} otherwise. The returned subset $R_n$ contains the layers that should be pruned from the model. Therefore, the goal of $\mathcal{A}$ is to \emph{find a subset $R_n$ of size $n$, so that the transformer model with layers $L \setminus R_n$ maximizes some -- previously specified -- performance measure $\mathcal{M}(L \setminus R_n, T)$ after being fine-tuned \footnote{\citeauthorn{poor-mans-bert} already showed that layer-wise pruning methods work best if executed directly after the pre-training, something we could confirm in our experiments.} on the task $T$}.

The most used layer-wise pruning algorithm is Top-layer Pruning.
\citeauthorn{poor-mans-bert} tested many different task-independent as well as task-dependent pruning algorithms and found that Top-layer pruning, which prunes the last layers first, works best across many different tasks and models:

\begin{equation}
    \prunetop(L, T, n) = \{L_d, L_{d-1}, \dots,  L_{d-(n-1)}\}
\end{equation}

Creating an algorithm $\pruneopt(L, T, n)$ that finds the optimal solution for layer-wise pruning is easy if resources are not constrained. More precisely, from all possible combinations to prune $n$ layers, we simply select the combination of layers that produces the highest performance:

\begin{equation}\label{eq:optimum}
    \pruneopt(L, T, n) = \argmax_{R_n}\{R_n \in \mathcal{P}_n(L) \mid \mathcal{M}(L \setminus R_n, T) \}
\end{equation}
where $\mathcal{P}(L)$ denote the set of all subsets of $L$ and $\mathcal{P}_n(L) = \{X \in \mathcal{P}(L) \mid |X| = n\}$. The computational complexity of this algorithm is therefore $\mathcal{O} {d \choose n}$.

For example, to prune two layers of a BERT model with 12 layers, 66 different networks must be evaluated. We show later that at least six layers must be pruned from models like BERT$_{\text{base}}$ or RoBERTa$_{\text{base}}$ to be competitive w.r.t memory consumption and inference time against state-of-the-art distillation methods. Pruning six layers already leads to 924 different configurations that must be evaluated. For the whole GLUE benchmark, this would take almost half a year using an NVIDIA 3090 GPU.

\subsection{Greedy-layer Pruning}
To reduce the complexity of the optimal solution, we exploit a \emph{greedy} approach. Greedy algorithms make a locally-optimal choice in the hope that this leads to a global optimum. For example, to prune $n+1$ layers, only a single (locally-optimal) layer to prune must be added to the already known solution for pruning $n$ layers. This statement would lead to the following assumption about layer-wise pruning:

\begin{assumption}[Locality]\label{asm:locality}
If the set $R_n$ ($R_0 = \emptyset$) is the optimal solution w.r.t. the performance metric $\mathcal{M}(L \setminus R_n, T)$ for pruning $n$ layers on a given task $T$ and model with layers $L$, then $R_n \subset R_{n+1}$.
\end{assumption}
The correctness of this assumption will be evaluated in an experimental study (\cref{sec:experiment_assumption}). Assuming locality, the search space and therefore the computational complexity is greatly reduced:
\begin{equation}\label{eq:greedy}
    \prunegreedy(L, T, n) =
    \begin{cases}
        \pruneopt(L, T, 1) & n = 1 \\
        \pruneopt\left(L \setminus R_{n-1}, T, 1\right) \bigcup R_{n-1} & \text{else}
    \end{cases}
\end{equation}
with $R_n = \prunegreedy(L, T, n)$.

It can be seen that $R_{n}$ can be calculated by adding the best local solution $\pruneopt\left(L \setminus R_{n-1}, T, 1\right)$ to the set $R_{n-1}$ using the locality \cref{asm:locality}. This simplification reduces the complexity from $\mathcal{O} {d \choose n}$ to $\mathcal{O}(n \times d)$. This algorithm can easily be distributed on multiple GPUs, e.g., if $d$ GPUs are available, the algorithm can be executed in just $n$ steps.

Our goal is to compress and train models with layer-wise pruning just before fine-tuning to (1) allow a fine-grained selection of the performance/speedup tradeoff and (2) to avoid the computationally expensive pre-training phase. It can be seen in \cref{eq:greedy} that the solution for pruning $n$ layers also includes the solution for pruning $n-1$ layers. Therefore, if $R_n$ is known, all other pruning combinations $R_x$ with $x \leq n$ can be determined in $O(1)$ and the performance/speedup trade-off can be selected individually without the need to run GLP or pre-training again.

\section{Experimental evaluation} \label{sec:experimens}
We first provide results on the GLUE benchmark, comparing Greedy-layer pruning to Top-layer pruning and different distillation methods. Next, we will evaluate the validity of the main components of GLP, the locality \cref{asm:locality} and the performance metric, $\mathcal{M}$.

\begin{table*}
\caption{Median of five runs evaluated for GLP, Top-layer pruning and state-of-the-art task-agnostic distillation methods. The subscript $n$ indicates how many layers are pruned from the given architecture. *denotes that the values are from the corresponding papers.} \label{tbl:glue}

\hspace{-0.5cm}
\begin{tabular}{l|cccccccc|cc}
Method &  SST-2 &  MNLI &  QNLI &   QQP (F1/Acc) &  STS-B &   RTE &  MRPC (F1/Acc) &  CoLA &  GLUE & Rel. \\
\hline
& &\multicolumn{6}{c}{Layer Pruning BERT} &\\
 Baseline$_0$ &  {92.8} &  {84.7} &  {91.5} &  88.0 / 91.0 &  {88.1} &  64.6 &  {88.7 / 83.6} &  56.5 &  {81.9} & - \\
      Top$_2$ &  92.3 &  83.9 &  89.8 &  87.4 / 90.5 &  88.0 &  64.3 &  87.1 / 81.4 &  {57.0} &  81.2 & 99.1\\
      GLP$_2$ (ours) &  92.3 &  83.9 &  90.7 &  87.5 / 90.7 &  87.6 &  65.3 &  87.8 / 82.6 &  {57.0} &  \textbf{81.5} & \textbf{99.5} \\
      Top$_4$ &  91.5 &  83.0 &  89.1 &  87.1 / 90.4 &  88.0 &  64.3 &  86.9 / 80.9 &  49.2 &  79.9 & 97.5 \\
      GLP$_4$ (ours) &  91.6 &  83.0 &  89.8 &  87.1 / 90.3 &  87.9 &  {66.1} &  85.9 / 79.9 &  54.5 &  \textbf{80.7} & \textbf{98.5}\\
      Top$_6$ &  90.8 &  81.1 &  87.4 &  86.7 / 90.0 &  87.7 &  63.5 &  86.1 / 79.4 &  37.1 &  77.6 & 94.7 \\
      GLP$_6$ (ours) &  90.7 &  81.2 &  87.7 &  86.3 / 89.7 &  87.6 &  60.6 &  85.4 / 77.9 &  45.0 &  \textbf{78.1} & \textbf{95.4}\\
 \hline
 & &\multicolumn{6}{c}{Layer Pruning RoBERTa} &\\
 Baseline$_0$ &  94.4 &  87.8 &  92.9 &  88.4 / 91.3 &  90.0 &  73.6 &  92.1 / 89.0 &  57.5 &  84.6 & - \\
      Top$_2$ &  94.3 &  87.7 &  92.5 &  88.4 / 91.3 &  89.9 &  69.7 &  90.9 / 87.3 &  58.1 &  \textbf{83.9} & \textbf{99.2} \\
      GLP$_2$ (ours) &  94.3 &  87.7 &  92.8 &  88.5 / 92.8 &  89.9 &  70.0 &  90.3 / 86.8 &  57.5 &  \textbf{83.9} & \textbf{99.2} \\
      Top$_4$ &  93.6 &  87.2 &  92.4 &  88.3 / 91.2 &  89.4 &  67.9 &  91.3 / 87.7 &  54.2 &  83.0 & 98.1 \\
      GLP$_4$ (ours) &  93.6 &  87.2 &  92.4 &  88.3 / 91.2 &  89.0 &  69.0 &  90.7  / 87.0 &  54.4 &  \textbf{83.1} & \textbf{98.2}\\
      Top$_6$ &  92.5 &  84.7 &  90.9 &  87.6 / 90.7 &  88.3 &  58.1 &  88.3 / 83.6 &  46.6 &  79.6 & 94.0 \\
      GLP$_6$ (ours) &  92.4 &  85.6 &  90.7 &  87.6 / 90.7 &  88.3  &  60.6 &  88.6 / 84.1 &  51.4 &  \textbf{80.6} & \textbf{95.3}\\
      LayerDrop$_6^*$ &  92.5 &  82.9 &  89.4 &  - &  - &  - &  85.3 / - &  - &  - & -\\
\hline
 & &\multicolumn{6}{c}{Task-agnostic Distillation} &\\
   DistilBERT &  90.1 &  82.3 &  88.8 &  86.6 / 89.9 &  86.3 &  60.6 &  88.9 / 84.1 &  49.4 &  79.1  & - \\
   MobileBERT & 91.9 &  {84.0} &  {91.0} &  87.5 / 90.5  &  {87.9} &  64.6 &  90.6 / 86.8 &  {50.5} &  {81.0} & - \\
   PKD* & {92.0} & 81.0 & 89.0 & 70.7 / 88.9 & - & {65.6} & 85.0 / 79.9 & - & - & - \\
   CODIR* & 93.6 & 82.8 & 90.4 & - / 89.1 & - & 65.6 & - / 89.4 & 53.6 & - & -\\

\end{tabular}
\end{table*}

\subsection{Setup} \label{sec:experimens_setup}

\paragraph{Implementation}
Each experiment is executed on a node with one Nvidia RTX 3090 GPU, an Intel Core i7-10700K CPU and 128GB of memory. Our publicly available implementation extends the BERT model as well as the RoBERTa model of the Transformer v.4.3.2 library (PyTorch 1.7.1) from HuggingFace with Top-layer pruning as well as Greedy-layer pruning. The \texttt{run\_glue.py} \footnote{\url{https://huggingface.co/transformers/v2.5.0/examples.html}} script from Huggingface is used to evaluate the GLUE benchmark. The source code also provides scripts to easily setup and reproduce all experiments.

\paragraph{Evaluation}
The GLUE benchmark \cite{glue} is designed to favor models that share general language knowledge by using datasets with limited training dataset size. It is a collection of 9 different datasets containing single sentence or sentence pair classification as well as a regression task, namely, CoLA \cite{cola}, SST-2 \cite{sst2}, MRPC \cite{mrpc}, STS-B \cite{stsb}, QQP \footnote{\url{http://qim.fs.quoracdn.net/quora_duplicate_questions.tsv}}, MNLI \cite{mnli}, QNLI \cite{qnli}, and RTE \cite{rte}. Following \citet{bert} we also exclude the problematic WNLI dataset from the benchmark \footnote{See also \url{https://gluebenchmark.com/faq}}.

To ensure that the layers that are selected by GLP for pruning also generalize and do not overfit the test-set, we use a 15\% split of the training set to compute $\mathcal{M}(L, T)$. Additionally, we report the median for \emph{five runs using different seeds for all our experiments} to get a fair comparison following \citeauthorn{distilbert}. We use the following classical performance metrics $\mathcal{M}$ \cite{glue}: the F1-score for MRPC and QQP, the Spearman correlation for STSB, the Mathews correlation for COLA and the accuracy for all other datasets. To ensure that our scores are not the result of any undesired side-effect or implementation details, we also executed experiments on Top-layer pruning and all distillation methods using exactly the same code-base, if a HuggingFace implementation was provided. Otherwise we report the values from the original paper.

\paragraph{Hyperparameters}
To get a similar speedup with pruning as with distilled models, the depth of a BERT model must be reduced to 6 layers (\cref{fig:timing}). We follow the hyperparameter setup from \citeauthorn{bert}, use a batch size of 32 and fine-tune the base model with a sequence length of 128. Although several optimizers have been applied in different work to obtain results with high performance, such as Lagrangian optimizer \cite{lagrangian-optimization} or Sobolev gradient based optimizer \cite{sobolev-grad-based-opt, capsnet_tumours}, they generally cause high computational costs. Therefore, we applied the AdamW optimizer \cite{adamw} in this work with a learning rate of 2e-5, $\beta_1=0.9$ and $\beta_2=0.999$.
To ensure reproducability of our results, we not only publish the source code together with the paper, but we also provide a guild\footnote{\url{https://guild.ai}} file (\texttt{guild.yml}) which contains the detailed hyperparameter setup we used for each experiment and which should help to replicate our experiments.

\subsection{Results on GLUE}
\begin{figure}
\centering
\includegraphics[width = 3.0in]{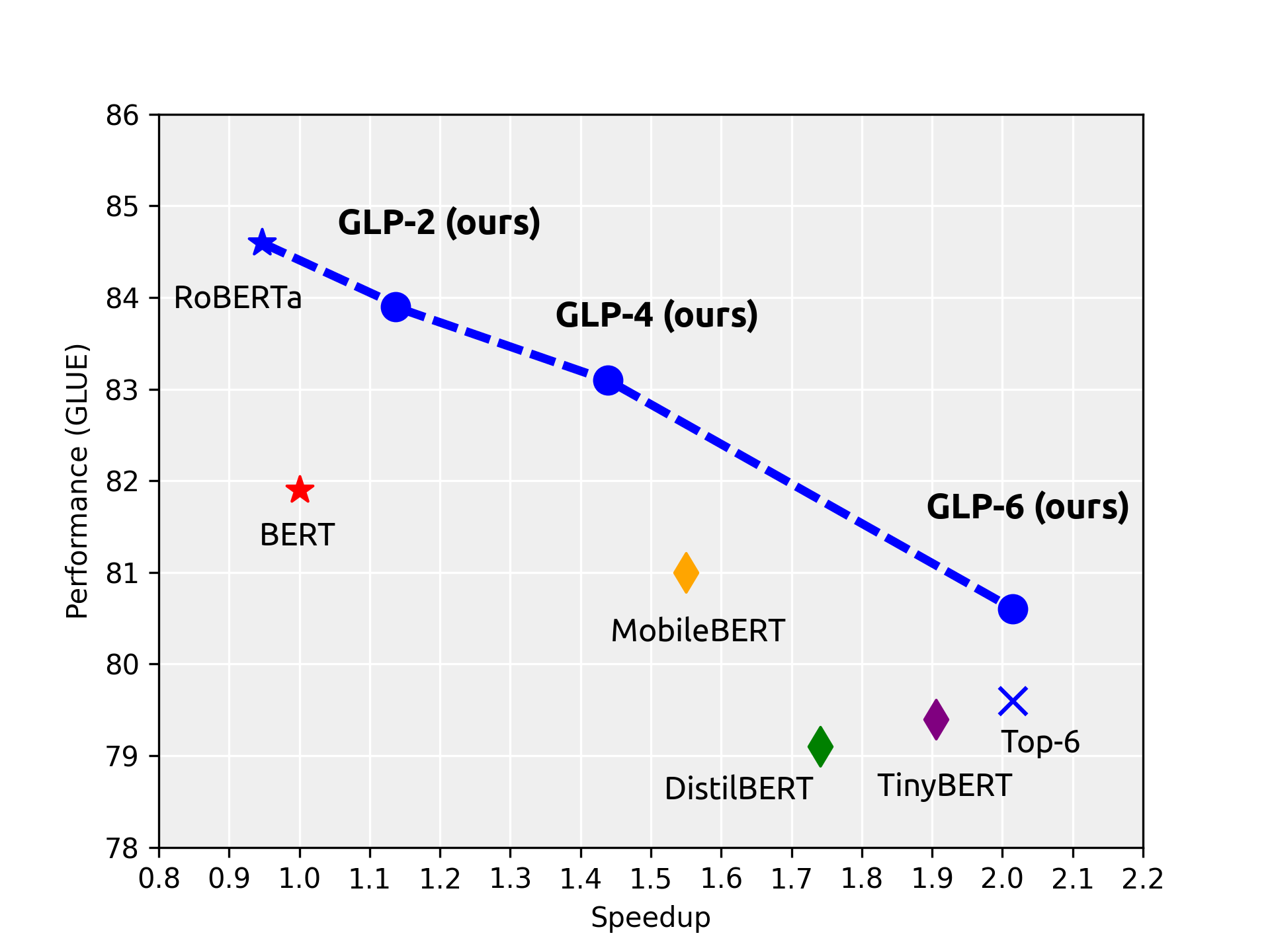}
\caption{Performance on the GLUE dataset (except for WNLI) using different models w.r.t its speedup relative to BERT. The dashed line shows how the performance/speedup ratio can be changed dynamically using GLP. In this specific case GLP was applied to a RoBERTa model, altough GLP is not restricted to this type of model.}
\label{fig:timing}
\end{figure}

Results for the baseline BERT and RoBERTa, as well as pruning 2, 4, and 6 layers from those models are shown in \cref{tbl:glue} and \cref{fig:timing} on GLUE. Greedy-layer pruning outperforms Top-layer pruning in almost all cases. For pruning 2 layers from RoBERTa, Top-layer pruning and GLP perform equally. Pruning 6-layer is particularly interesting as it allows a comparison to distillation methods w.r.t performance as well as inference time. In this case, GLP always outperforms Top-layer pruning, sometimes by a large margin: The difference in performance is 0.5 for pruning BERT and 1.0 percentage points for pruning RoBERTa. When compared against LayerDrop -- which is a fairly expensive approach because it must be executed during both pre-training and finetuning -- GLP outperforms it or reaches similar scores.

On some tasks, GLP and Top-layer pruning perform equally well. For example, when pruning 2 layers from BERT on CoLA, the score is exactly the same. As an interesting point, we found that in this case, GLP prunes exactly the same layers as Top-layer pruning. But, in most of the other cases, solutions are quite different, which indicates that the optimization problem (\cref{sec:method}) is not task-independent. For example, the layer-pruning sequence when using GLP for BERT on CoLA is $\{11,10,9,4,0,5\}$ with GLP, which is quite different when compared against Top-layer pruning (sequence $\{11,10,9,8,7,6\}$). The accuracy of GLP$_6$ is also much higher with 45.0, compare to Top$_6$ with only 37.1. Note that we provide all layers found by GLP for each task and model together with our source code to support future research and reproducibility.

In \cref{tbl:glue} (column Rel.) we additionally show the relative performance compared to its baseline. The relative performance of Greedy-layer pruning w.r.t the baseline is not only higher in almost all cases than Top-layer pruning, but results are also more consistent for different types of models. For example, the difference of the relative performance between BERT and RoBERTa is only $0.1\%$ for GLP$_6$, but $0.7\%$ for Top$_6$. We believe that one major reason is that our algorithm selects layers based on the given task and model i.e. is task-dependent whereas Top-layer pruning is task-independent. This strongly indicates that the proposed Greedy-layer pruning algorithm can also be used for models that will be released in the future so that even better results can be achieved by simply applying GLP to those pre-trained models. It is worth mentioning that the relative performance can be larger than $100\%$ for some datasets. For example, when using GLP to prune 2 or 4 layers from BERT, the resulting model trained on RTE is not only faster and smaller but also outperforms the baseline. A hypothesis for this phenomenon is given later in the discussion \cref{sec:conclusion}.

\subsection{Comparison against Knowledge Distillation}

\begin{table*}
\centering
\caption{Comparison of base models, state-of-the-art layer-wise pruning and distillation models w.r.t inference time, memory consumption and performance on GLUE. *denotes that this compression method uses a teacher model in the fine-tuning stage whereas all other models directly train on the downstream task  \cite{mobilebert}}\label{tbl:distillation}
\begin{tabular}{ll|cccc}
 &  &   & & \multicolumn{2}{c}{Inference [s]} \\
 Method & Model &  GLUE & Params. & CPU & GPU \\
\hline
Base & BERT & 81.9 & 108M & 141 & 3.0 \\
 & RoBERTa & \textbf{84.6} & 125M & 149 & 3.0 \\
\hline
Task-agnostic distillation & DistilBERT & 79.1 & 67M & 81 & \textbf{1.7} \\
& MobileBERT & 81.0 & \textbf{25M} & 91 & 9.1 \\
\hline
Non task-agnostic distillation & TinyBERT$_6^*$ & 79.4 & 67M & 74 & \textbf{1.7} \\
\hline
Layer-wise pruning  & LayerDrop$_6$ & - & 82M & \textbf{70} & \textbf{1.7} \\
& BERT+Top$_6$ & 77.6 & 67M & 73 & \textbf{1.7} \\
 & BERT+GLP$_6$ (ours) & 78.4 & 67M & 73 & \textbf{1.7} \\
 & RoBERTa+Top$_6$ &  79.6 & 82M & \textbf{70} & \textbf{1.7}\\
 & RoBERTa+GLP$_6$ (ours) & 80.6 & 82M & \textbf{70} & \textbf{1.7}
\end{tabular}
\end{table*}

Memory consumption, inference time, and performance on GLUE for baseline models, distillation models, and pruning methods are shown in \cref{tbl:distillation}. No GLUE scores were reported for LayerDrop in \citeauthorn{structured-dropout}. Performance and speedup values relative to BERT are additionally shown in \cref{fig:timing}. To measure the inference time, we evaluated each model on the same hardware, using the same frameworks with a batch-size of $1$ and perform the evaluation on a CPU following \citeauthorn{distilbert}. Additionally, for each method we report the GPU timings, measured on an RTX 3090 GPU with a batch size of 32.

In terms of inference-time and performance (\cref{fig:timing}), GLP is indeed on-par when compared to  distillation methods. For example it outperforms TinyBert as shown in \cref{fig:timing} w.r.t. performance and inference time. GLP outperforms PKD \cite{pkd} in 5 out of 6 cases and outperforms CODIR \cite{codir} in 3 out of 7 cases. Therefore, it can be concluded that GLP is on-par with state-of-the-art knowledge distillation methods. Additionally, pruning allows for balancing the speed-up/performance trade-off by changing the number of layers that should be pruned from the model (dashed line in \cref{fig:timing}) which is another huge advantage as we motivated already in \cref{sec:introduction}. In terms of memory consumption and  performance, GLP is very close to DistilBERT and TinyBERT, but the winner here is MobileBERT, which outperforms all methods. MobileBERT was introduced especially for mobile devices and so is extremely parameter efficient as it is a thin and deep architecture. On the other hand, inference of MobileBERT is slow on GPUs because parallelism on GPUs can be barely  exploited with such an architecture. Nevertheless, this aspect shows  the limitation of layer-wise pruning in that the width is fixed and only the depth of a network can be changed. More insights on this will be discussed in \cref{sec:conclusion}.

Executing GLP on an average downstream-task of GLUE takes about 20h on a single 3090 GPU on the QNLI task. A similar setup in the cloud would be a single Tesla P100 GPU which costs approximately $32\$$ \footnote{Pricing from \url{https://cloud.google.com/compute/gpus-pricing}. Accessed 07/09/2021}. After the execution of GLP arbitrary performance / speedup ratios can be selected by the user through pruning a certain number of layers, without the need of an additional pre-training phase.

On the other hand, to select an individual performance / speedup ratio with knowledge-distillation, a new student-model must be pre-trained from scratch and fine-tuned on the downstream task. Therefore, an additional and computationally expensive pre-training phase is necessary. A financially cheap methodology to pre-train models is presented by \citeauthorn{academic-budget} and would cost about $400\$$ if executed on the proposed hardware setup which is $12 \times$ more expensive than the execution of GLP.

\subsection{Evaluating the Locality \Cref{asm:locality}} \label{sec:experiment_assumption}
In this subsection we evaluate the two main components of GLP, namely, the effect of the locality \cref{asm:locality} and the influence of the performance metric $\mathcal{M}$ on the results.

\begin{figure*}
\centering
\begin{tabular}{cc}
\subcaptionbox{BERT\label{fig:optimal_bert}}{\includegraphics[width = 4.6in]{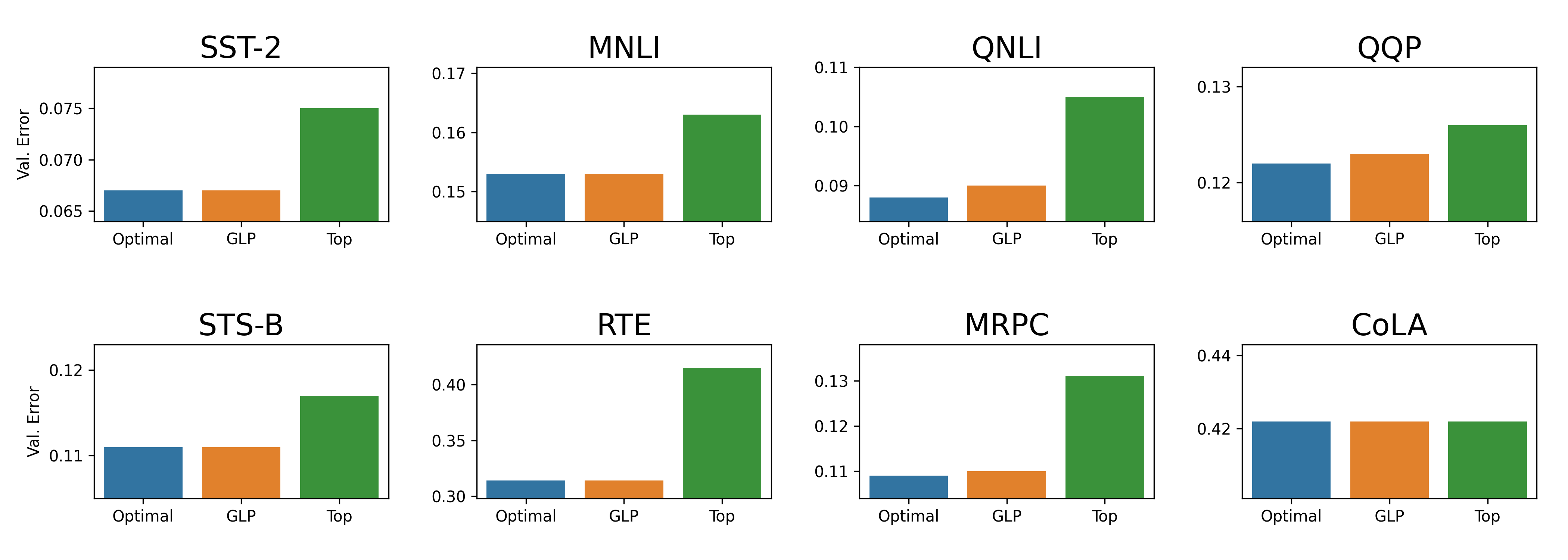}} \\
\subcaptionbox{RoBERTa\label{fig:optimal_roberta}}{\includegraphics[width = 4.6in]{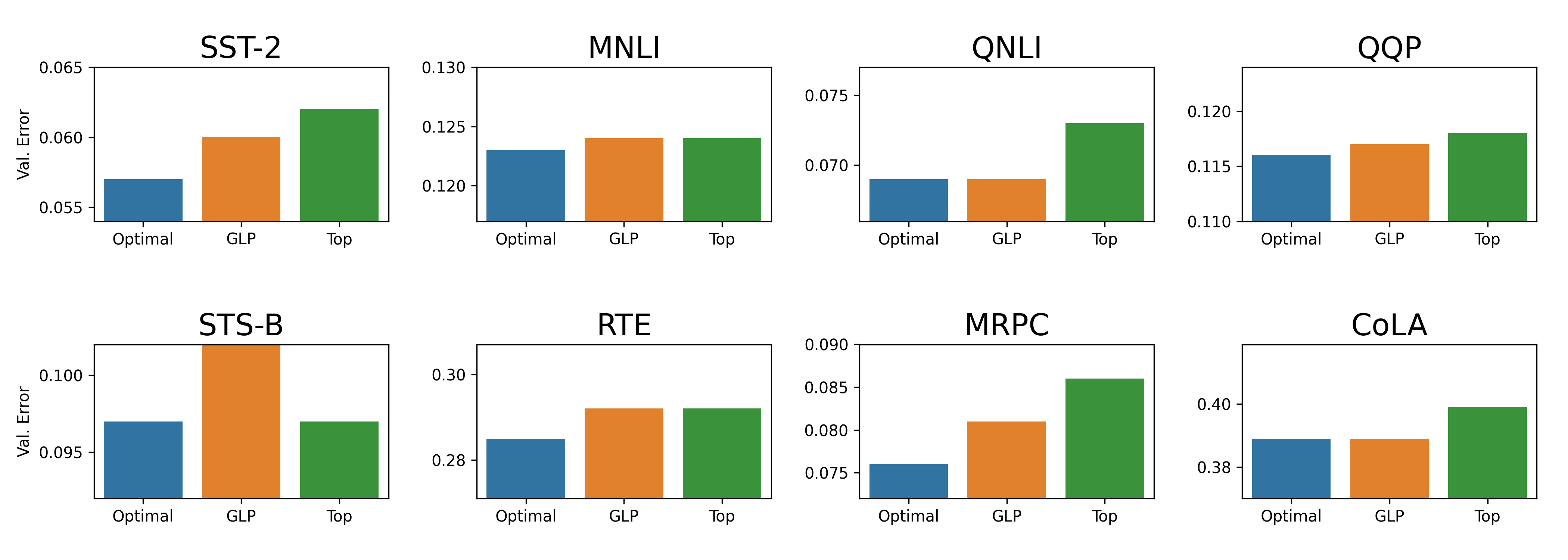}} \\
\end{tabular}
\caption{Comparison of the optimal solution and the solutions found by Top-layer pruning and greedy pruning for removing two-layers from a BERT model and a RoBERTa model. We executed it for all datasets of GLUE and present the results for the first datasets in those graphs.}
\label{fig:optimal}
\end{figure*}

We assess first if a locally optimal choice also leads to a globally optimal solution (\cref{asm:locality}) by evaluating all of the possible two-layer combinations pruning from a transformer model. The \emph{Optimal} solution is shown in \cref{fig:optimal}, next to GLP and Top-layer pruning. If \cref{asm:locality} were to be correct, GLP should find a solution with the same validation error for each dataset and model. Results for the optimal solution (\emph{Optimal} in \cref{fig:optimal}) are computed on a fixed seed as it is important to use the same seed in order to properly evaluate if the solutions for GLP and Optimal are exactly the same. In all other experiments we used multiple runs with different seeds and report the median to exclude outliers and to ensure that GLP does not depend on a single seed. We also want to clarify that the computational resources to compute the optimal solution grows exponentially with the number of layers that should be pruned. Unfortunately, pruning of more than two layers becomes computationally unfeasible and therefore values for pruning two layer are reported.

\Cref{fig:optimal} shows that solutions found by GLP almost always outperform Top-layer pruning by a large margin, which indicates the robustness of task-dependent algorithms. The exception in this case is for RoBERTA on STS-B , where the solution found with GLP is slightly worse than Top-layer pruning. GLP can find the optimal solution in many cases, for example, when applied to BERT (\cref{fig:optimal_bert}), that is the case for SST-2, RTE, STS-B, MNLI and CoLA. In others, even though GLP does not provide such optimal solution it is a close one (such as QQP, MRPC and QNLI for BERT). \Cref{fig:optimal_roberta} provides the same evaluation for RoBERTa. This detailed analysis proves that the locality assumption is an (although very good) approximation. Even so, the assumption is useful because it drastically reduces the size of the search space which grows exponentially with the number of layers that should be pruned as we already described in \cref{sec:method}.

Finally, we can also evaluate the choice for the performance metric $\mathcal{M}$ used by the GLP algorithm (\cref{eq:greedy}) to find locally-optimal solutions. More precisely, we replaced $\mathcal{M}$ with the validation loss and pruned layers that produce the lowest loss value. We found in our experiments that this setup worsened the overall performance of GLP on the GLUE benchmark. For example, for pruning a BERT model, the score is worsened on average by $0.7$ percentage points.

\section{Discussion and Conclusions} \label{sec:conclusion}
In this paper, Greedy-layer pruning is introduced to compress transformer models dynamically just before the fine-tuning stage to allow a fine-grained selection of the performance/speedup tradeoff. Our focus is on pruning layers just before the fine-tuning stage as the pre-training stage is computationally expensive. Therefore, the execution of GLP is $12 \times$ less expensive than state-of-the-art knowledge distillation methods. The GLP algorithm presented here implements a greedy approach maintaining very high performance while needing only a modest budget computational setup. This is empirically evaluated in \cref{sec:experimens} by comparing GLP against state-of-the-art pruning and distillation methods. An interesting result is that GLP sometimes improved the performance of its baseline (\cref{tbl:glue}) although it is a smaller and faster model. This phenomenon also occurs for other types of models as already shown before \cite{limitation-capsule} and could be the result of layers that exist in neural networks that worsen the accuracy of the trained model \cite{conflicting-bundles}. In this sense, GLP acts as a neural architecture search algorithm that prunes layers that worsen the accuracy first. Therefore, the performance of the pruned model can be larger than the performance of its baseline.

Some \emph{limitations} of the proposed methodology include that GLP changes only the depth of networks by pruning layers from transformer models, but it is not possible to reduce the width of models to further compress models. Additionally, GLP is task-dependent and therefore pruning must be executed with the task-specific dataset. Finally, for very deep networks with more than 50 layers (such as CNNs) the proposed greedy strategy becomes computationally expensive, and more optimization strategies would be needed.

\section{Future Work} \label{sec:future_work}
We have shown that the locality \cref{asm:locality} is a good approximation to find comparable solutions to expensive distillation systems, but with the advantage of doing so when only limited computational resources are available. Nevertheless, there is still room for improvement w.r.t performance as shown in our experimental evaluation (\cref{fig:optimal}). One interesting direction to further improve the performance could be to evaluate other different methodologies than the ones presented in the greedy approach. Secondly, the performance metric $\mathcal{M}$ used by GLP for evaluation could be a topic for further research. For example, \citet{academic-budget} rejects training candidates already after a few training steps if a given threshold is not reached. A similar approach for GLP could be a topic of interest to speed up the compression stage and decrease computational costs even further.

\bibliographystyle{model2-names}
\bibliography{refs}

\end{document}